\documentclass[12pt, letterpaper]{article}
\usepackage[letterpaper, margin=1.5in]{geometry}
\usepackage[utf8]{inputenc}
\usepackage{lipsum}
\usepackage{amsmath,amssymb,amsthm}
\usepackage{soul,xcolor}
\usepackage{hyperref}
\usepackage{todonotes}
\usepackage[onehalfspacing]{setspace}   

\definecolor{bittersweet}{rgb}{1.0, 0.44, 0.37}
\definecolor{babyblue}{rgb}{0.54, 0.81, 0.94}
\definecolor{bananayellow}{rgb}{1.0, 0.88, 0.21}

\usepackage{graphicx, color}

\title{{Attribution-based XAI Methods in\\Computer Vision: A Review}}
\author{Kumar Abhishek\textsuperscript{\textdagger}, Deeksha Kamath\textsuperscript{\textdagger}\\
{School of Computing Science, Simon Fraser University, Canada}\\
\texttt{\{kabhishe, dkamath\}@sfu.ca}
\thanks{\textsuperscript{\textdagger}Equal contribution. Authors ordered alphabetically. Work done in Fall 2020.}
}

\date{}

\renewcommand\footnotemark{}

\begin{document}

\maketitle

\begin{abstract}

The advancements in deep learning-based methods for visual perception tasks have seen astounding growth in the last decade, with widespread adoption in a plethora of application areas from autonomous driving to clinical decision support systems. Despite their impressive performance, these deep learning-based models remain fairly opaque in their decision-making process, making their deployment in human-critical tasks a risky endeavor. This in turn makes understanding the decisions made by these models crucial for their reliable deployment. Explainable AI (XAI) methods attempt to address this by offering explanations for such black-box deep learning methods. In this paper, we provide a comprehensive survey of attribution-based XAI methods in computer vision and review the existing literature for gradient-based, perturbation-based, and contrastive methods for XAI, and provide insights on the key challenges in developing and evaluating robust XAI methods.
\end{abstract}

\section{Introduction}

The past decade has witnessed unprecedented advancements in the development of artificial intelligence algorithms, and in particular, neural networks for deep representation learning-based methods applied to visual perception tasks~\cite{schmidhuber2015deep,lecun2015deep}. This growth has, in large part, been fueled by the advent of large public datasets~\cite{deng2009imagenet,geiger2012we,lin2014microsoft,everingham2015pascal,cordts2016cityscapes,simpson2019large}, optimal utilization of computational resources~\cite{raina2009large,nickolls2010gpu,konevcny2016federated}, and efficient optimization methods~\cite{duchi2011adaptive,kingma2014adam}. Owing to their parameter-efficient architectures and the ability to exploit spatial structure information, convolutional neural networks (CNNs)~\cite{krizhevsky2012imagenet,zeiler2014visualizing,simonyan2014very,szegedy2015going,he2016deep} have consequently become the backbone of many computer vision applications and are now ubiquitous in a wide range of applications, from autonomous driving vehicles~\cite{grigorescu2020survey} to clinical decision support systems for automated diagnosis of disease conditions with near-expert level accuracies~\cite{gulshan2016development,esteva2017dermatologist,rajpurkar2017chexnet}. However, despite their superior performance, these models often tend to be opaque in terms of their operation, acting as black-box models with little to no insight to their decision making process.

Initially coined by van Lent et al.~\cite{van2004explainable} to explain decisions made by artifical intelligence systems in simulations, XAI (eXplainable AI) methods attempt to explain machine learning models and the decisions made by them. In their highly cited paper, Doshi-Velez et al.~\cite{doshi2017towards} highlighted the need to formalize the definition of interpretability in machine learning and find discrete ways of evaluating them, and proposed a taxonomy of interpretability evaluation methods. Soon after, Samek et al.~\cite{samek2017explainable} discussed the importance of interpretability of deep learning-based AI system and the impact of an incorrect understanding of their underlying workings. They presented four arguments in favor of XAI methods: verification of these systems, improvement of these systems, the ability to transfer a system's learned knowledge to humans, and legislative compliance of such systems. The latter is especially important with the recent adoption of the \textit{General Data Protection Regulation} (GDPR) in 2018, with a requirement that machine learning algorithms practice efficiency along with fairness and transparency as well as a ``right to explanation" for the development and deployment of automated data analysis systems~\cite{goodman2016european}. While the concept of model interpretability gained widespread recognition in academia and the general public, a myriad of definitions of model interpretability also came into existence, with no single unversally agreed upon definition~\cite{lipton2018mythos}.

Given the need for explaining deep learning-based prediction models, there have been previous surveys on this topic. Chakraborty et al.~\cite{chakraborty2017interpretability} categorized the prior work on machine learning interpretability along two dimensions: model transparency and model functionality, and discussed the literature at high level without much details. They stated that model fairness and accountability are the key challenges, and that a Bayesian approach to deep learning may address some of these challenges. Guidotti et al.~\cite{guidotti2018survey} and Carvalho et al.~\cite{carvalho2019machine} provided a comprehensive survey of methods for explaining machine learning methods including a taxonomy of model explainability approaches without any particular focus on CNNs. Zhang et al.~\cite{zhang2018visual} reviewed the literature for understanding neural network representations with a specific emphasis on disentangling the latent space representations learned by CNNs. Du et al.~\cite{du2019techniques} conducted a survey of post-hoc interpretability methods and categorized previous works into global and local explanations as well as model-agnostic and model-specific explanations. Borji~\cite{borji2019saliency} presented a comprehensive survey of the literature in predicting visual saliency for deep learning models, including a review of evaluation metrics, performance benchmarks, and open problems, but, did not cover any attribution-based XAI methods. The closest survey to our review is a very recent paper by Singh et al.~\cite{singh2020explainable} which covered the literature for XAI methods for deep learning, but focused only on medical images. Moreover, the majority of the methods covered belonged to gradient-based approaches and only two perturbation-based approaches.

In this paper, we survey the literature for explainability methods for deep learning-based computer vision models. Unlike existing surveys which are either too broad in the areas they cover~\cite{guidotti2018survey,carvalho2019machine} or cover only a  sparse collection of the literature~\cite{du2019techniques} with a focus on a certain kind of images (i.e., medical images~\cite{singh2020explainable}), our survey focuses on attribution-based XAI methods for explaining the predictions of deep CNNs for general computer vision, thus striving for the perfect balance between the breadth and the depth of material reviewed. 

\section{Attribution-based XAI Methods}

\begin{figure}[!tbp]
    \centering
    \includegraphics[width=\textwidth]{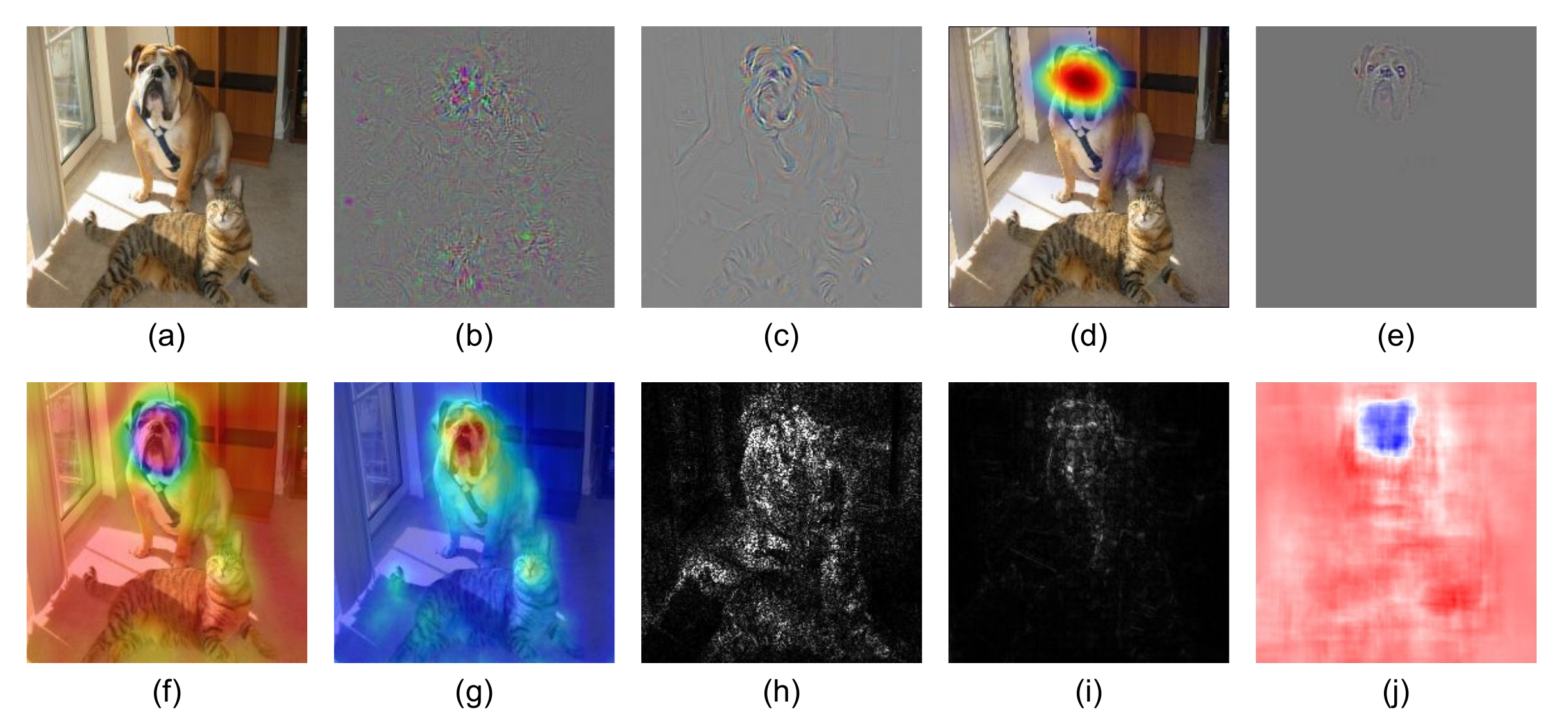}
    \caption{Visualizing attribution-based XAI methods' outputs for `dog' class: (a) original input image, (b) backpropagated gradients, (c) guided backpropagation, (d) Grad-CAM, (e) guided Grad-CAM, (f) Score-CAM, (g) FullGrad, (h) Integrated Gradients, (i) SmoothGrad, and (j) occlusion sensitivity map.}
    \label{fig:examples}
\end{figure}

Consider a deep neural network model parametrized by $\mathcal{F}: \mathcal{X} \to \mathcal{Y}$, where $x = [x_1, \cdots, x_{D_{in}}] \in \mathbb{R}^{D_{in}}$ is a ${D_{in}}$-dimensional input to the model, which produces $y = [y_1, \cdots, y_{D_{out}}] \in \mathbb{R}^{D_{out}}$, a ${D_{out}}$-dimensional output. While working with images, the input dimensionality can get quite large, with $D_{in}$ being more than 150,000 for a $224 \times 224$ RGB image. In order to explain model predictions, we seek to assign attributions to each input feature (pixels in the case of image input) so as to understand their individual contributions to the model's prediction. For the $j^{\textrm{th}}$ unit in the model's output, an attribution method generates the contribution of each input feature $x_i \in x$ to the output $y$ as $\mathcal{A}^j = [\mathcal{A}^j_1, \cdots, \mathcal{A}^j_{D_{in}}] \in \mathbb{R}^{D_{in}}$~\cite{ancona2019gradient}. For an image classification model, we generate a single label for an input image, and therefore ${D_{out}} = 1$. The attribution therefore becomes the contribution of each spatial location in the input (pixel) towards the (in)correctly predicted class label. Because images have an inherent spatial structure, these attributions are typically visualized as heatmaps drawn on top of the input image to depict the contribution of the input features to the model predictions. In this survey, we cover three major categories of attribution-based methods: gradient-based, perturbation-based, and contrastive methods. Figure~\ref{fig:examples} shows the attribution maps for some of these methods applied to an image from ImageNet~\cite{deng2009imagenet,kazuto10112017smoothgrad,utkuozbulak2020convolutional}. We choose this image because of the presence of several confounding factors such as non-uniform background, illumination, shadow, and multiple objects.

\subsection{Gradient-based Methods}

Operating on the assumption that the parameters of the predictive models (i.e., the deep neural networks) are accessible, gradient-based methods exploit the information flow pathway in deep neural networks and use one (or a few) forward and backward pass to first infer the class label (in the forward pass) and then use the predicted label to back-propagate through the network sequentially upto the input layer to estimate the input attributions (in the backward pass). Gradient-based methods generate ``saliency maps", which indicate the contributions of each variable in the input space to the network's final prediction~\cite{ancona2019gradient}. An added advantage of gradient-based methods is that a single (or a few) forward and backward pass yields the importance scores of all the features in the input.

One of the earliest successful development of an explanation visualization technique was the deconvolutional network (DeconvNet) architecture proposed by Zeiler et al.~\cite{zeiler2014visualizing}, which aimed to approximately visualize the input stimuli that excite individual feature maps at each convolutional layer. A ReLU non-linearity is applied to the gradient computation to ensure a positive reconstruction of the image and in the subsequent max pooling operation, the locations of the maxima of each region were recorded. This allows identification of parts of an image responsible for the activation. The model decomposes an image in a hierarchical fashion using multiple alternating layers of deconvolution and un-pooling, and the cost function comprised of (a) a likelihood term that keeps the reconstruction of the input close to the original input image and (b) a regularization term that penalizes the $l_1$ norm of the 2D feature maps~\cite{zeiler2011adaptive}.

The novel DeconvNet, due to its dependence on ReLU nonlinearity was thus limited to CNN models with ReLU activation. Nevertheless, DeconvNets inspired Simoyan et al.~\cite{simonyan2013deep} to develop the saliency map, which is a generalization of the DeconvNet, as gradient-based techniques can be applied to any layer, including the convolutional layer. This corresponds to the visualisation of a neuron in a fully connected layer instead of a convolutional one. This method leverages the absolute value of the partial derivative of the target output neuron with respect to the input features to find the features of an image that affect the output the most with minimal perturbation. A major drawback that ensued was that the use of absolute values did not allow distinction between positive and negative derivative values.

Building on the approaches of DeconvNets and saliency maps, Springenberg et al.~\cite{springenberg2014striving} proposed the guided backpropagation algorithm, which included an additional guidance signal from the higher layers in addition to the usual backpropagation, hence `guided backpropagation'. They synthesized a CNN in which the max-pooling operation was replaced by a convolutional layer. In this technique, ReLU is applied to computation of gradient in addition to the gradient of a neuron with ReLU activation. Unlike the DeconvNet, guided backpropagation works well without switches, and provides the ability to visualize intermediate layers as well as the last layers of the network. 

Along the same timeline, as interactive visualization gained more traction, Yosinski et al.~\cite{yosinski2015understanding} developed a software tool that provided a live, interactive visualization of every neuron in a trained CNN as it responded to a user-provided image or video input. Backward derivatives are computed via backpropagation or deconvolution~\cite{zeiler2014visualizing} starting from arbitrary units, and the perceptual loss function is based on differences between high-level image feature representations extracted from pretrained CNNs. 

Along the same approach of generating saliency maps, Lin et al.~\cite{lin2013network} introduced an interpretable method called global average pooling (GAP) where they directly output the spatial average of the feature maps from the last `mlpconv' layer (a multi-layer perceptron which maps an input local patch to an output feature vector) as the confidence of categories via a global average pooling layer and the resulting vector is fed into the softmax layer. GAP acts as a structural regularizer by summing out the spatial information, thus preventing overfitting during training and making the model more robust to spatial translations of the input.

Due to the zero-ing out of negative gradients, both guided backpropagation and deconvolutional networks can fail to highlight inputs that contribute negatively to the output. As an alternative method, Zhou et al.~\cite{zhou2016learning} developed the class activation map (CAM). In their experiments, they discovered that the GAP layer not only served as a structural regularizer, but also enabled the network to retain its localization ability until the final layer, opening up possibilities of identifying discriminative regions in a single forward pass and generating class specific feature maps. A class activation map is the weighted activation map generated for each image, and when generated for a specific class, indicates the discriminative image regions employed by the CNN to identify that class.

A drawback of CAMs is that they require the GAP layer to immediately follow the convolutional layer being visualized. In the absence of a GAP layer or one (or more) fully connected layer(s), CAMs will fail due to the class-wise weights for each activation unit not being defined.
As a solution to this problem, Selvaraju et al.~\cite{selvaraju2017grad} introduced Grad-CAM, where they combined feature maps using the gradient signal. They achieved localization in one pass, requiring a single forward and a partial backward pass per image. The mean of the gradients of the logits of the class with respect to the feature activation maps of the final convolutional layer is computed, giving an importance score of a neuron.
In order to obtain features that have a positive influence on the class of interest,a ReLU nonlinearity is applied to to the weighted combination of forward activation map.

As Grad-CAM produces only coarse-grained visualization, the authors also combine guided-backpropagation~\cite{springenberg2014striving} with their method and propose a new method called Guided Grad-CAM~\cite{selvaraju2017grad}. They extract the element-wise product of guided-backpropagation visualization and Grad-CAM’s visualization, resulting in a visualization that is both fine-grained and class-discriminative. This method nonetheless also acquires the limitations of guided backpropagation caused by cancelling out negative gradients during backpropagation.

Building on Grad-CAM, Chattopadhyay et al.~\cite{chattopadhay2018grad} proposed a generalized version of the same, called Grad-CAM$++$, which significantly outperformed the previous methods in providing explanations for images with multiple occurrences of the same object, and could also be applied to spatio-temporal data. Using a single backward pass like previous gradient-based methods, Grad-CAM++ averages the pixel-wise weights of the gradients for obtaining higher-order derivatives for exponential and softmax activation functions. 
 
In order to address the limitations of Grad-CAM and Grad-CAM++, Wang et al.~\cite{wang2020score} developed a visual explanation technique called Score-CAM, where the importance of activation maps are encoded by the global contribution of the corresponding input features instead of the gradient information. The authors propose a new performance evaluation measure, Channel-wise Increase of Confidence (CIC), to attach an importance score to each activation map.

Sundarajan et al.~\cite{sundararajan2017axiomatic} developed the Integrated Gradients (IG) method, which determines the salient inputs by gradually varying the network input from a baseline to a current input at hand and iteratively aggregating the gradients along this path.

A challenge with all of the methods described above is that discontinuity in the gradients can produce misleading importance scores. Morover, these methods had an inability to identify negative contribution. Shrikumar et al.~\cite{shrikumar2017learning} proposed a different method that solved these challenges, DeepLIFT (Deep Learning Important FeaTures), which takes into consideration a `reference' output, and compares this reference to the actual output. The reference output can also represent a neutral input or the lack of an object previously present. DeepLIFT computes the reference values of all hidden units using a forward pass and then proceeds backwards, allowing information propagation even when gradient is zero. It has two variants, namely the Re-scale rule and RevealCancel which treats positive and negative contributions to a neuron separately, filling in a major gap present in the previous methods.

In recent times, Smilkov et al.~\cite{smilkov2017smoothgrad} described how gradient-based methods suffer from rapidly fluctuating partial derivatives and proposed SmoothGrad, a technique for smoothing out the gradients using a Gaussian kernel, to produce saliency maps with reduced noise and more coherent visualization. They further employ a stochastic approximation to calculate the local average of the class activation function.

More recently, Srinivas et al.~\cite{srinivas2019full} developed the FullGrad, a method to assign dual importance scores, one to input features and another for individual neurons within the network. This method therefore accounts for both local and global pixel importances, i.e., corresponding to both the individual input pixels as well as to the groups of input pixels.

\subsection{Perturbation-based Methods}

Unlike the gradient-based attribution methods which quantify the contribution of input features to the network's predictions, perturbation-based methods have a somewhat different objective: explaining how the network's predictions change when one or more input features (pixels) are perturbed, make perturbation-based methods an optimal choice for sensitivity analysis of the models~\cite{ancona2019gradient}. The sensitivity analysis of deep neural networks becomes especially important when dealing with carefully engineered perturbations, called ``adversarial perturbations"~\cite{szegedy2013intriguing} to the input which are imperceptible to the human eye but can cause catastrophic prediction failures. Unlike gradient-based methods which require access to the model parameters for calculating the gradients, perturbation-based methods only need to perform one or more forward passes. However, they are often computationally more expensive since they only estimate the importance of a subset of input features, thus requiring multiple inference calls~\cite{zintgraf2017visualizing}. Moreover, given the non-linear nature of deep neural network models, the outputs of these methods are specific to ``the exact subset and modification" of the input features, and are not generalizable to all possible perturbations~\cite{kapishnikov2019xrai}.

Occlusion sensitivity maps proposed by Zeiler et al.~\cite{zeiler2014visualizing} were one of the earliest perturbation-based model explanations. They studied a deep model's reasoning by repeatedly replacing different patches of the input image with a solid grey colored square and observing the predicted class label. They investigate if the model's prediction truly aligns with the object location in the image or if the model is simply using the surrounding pixels for context, and visualizations of the last convolutional layer's activation maps show that the model's predicted probability for the correct class label degrades progressively as more object area is occluded.

Such perturbation-based explanations for prediction models are not restricted to deep models only, and Ribeiro et al.~\cite{ribeiro2016should} proposed an explanation technique applicable to all black-box prediction models. Termed LIME (Local Interpretable Model-agnostic Explanations), the proposed method generates an interpretable model for any prediction model, and this interpretable model is locally faithful to the prediction model. Given a prediction model, the inputs are perturbed (in the case of images, by occluding certain regions of the image) to generate data samples in the neighborhood of the original data points, which are then weighed based on their proximity to the original data point in the input space and used to train a linear interpretable model. The explanations are then produced in the form of a coarse heatmap of the same size as the input and is denoted using fixed superpixels.

Building upon a previous approach for explaining model predictions by decomposing individual input feature contributions~\cite{robnik2008explaining}, Zintgraf et al.~\cite{zintgraf2017visualizing} proposed a ``probabilistically sound" approach for explaining the predictions of deep neural network models. They use conditional sampling and multivariate analysis to estimate the ``relevance" of a patch in the input image by calculating the difference in the model's predictions with and without the input patch, and this is simulated by marginalizing the input features. While the proposed approach is more rigorous in evaluating the effect of input feature removal as compared to Zeiler et al.~\cite{zeiler2014visualizing}, it is extremely slow, with explanations for a VGG-16 network~\cite{simonyan2014very} taking 70 minutes on average even when using GPU implementation.

Fong et al.~\cite{fong2017interpretable} introduced a model-agnostic image-editing-based explanation approach where the proposed method optimizes for a mask to explain the network's predictions for a certain class label. Starting with a low-resolution mask, the authors aim to produce a minimal and smooth mask which when used to perturb the input image (by using the input image, a blurred version of the input image, and an upsampled version of the learned mask) causes the score of that class label to drop considerably. The sparsity and the smoothness of the mask are enforced using an $\mathcal{L}_1$ constraint and a total variation regularization term in the optimization objective. In a follow-up work, Fong et al.~\cite{fong2019understanding} generalize this by introducing ``extremal perturbations", referring to a perturbation which has the maximum effect on the network's predictions for a class, among all perturbations of a given area. Instead of an explicit smoothness constraint, the perturbations are chosen from a parametric family with a minimum guaranteed smoothness. Finally, in addition to spatial masks, the proposed method also identifies salient channels by learning channel-wise attribution masks.

More recently, Petsiuk et al.~\cite{petsiuk2018rise} proposed a very simple black-box explanation technique called RISE (Randomized Input Sampling for Explanation) for generating pixelwise saliency maps. Given an input image and a prediction model, RISE feeds multiple randomly masked versions of the image to the model. The resulting model prediction scores for a certain class are then used as weights to generate a linear combination of the random masks, which yields the saliency map for that particular class. To generate these masked inputs, the input image is multiplied elementwise with random binary masks obtained using Monte Carlo sampling.

To summarize, while perturbation-based XAI methods are superior in that they do not need access to the model's parameters and only need forward passes through the model. However, they are often computationally more expensive since they only estimate the importance of a subset of input features, thus requiring multiple inference calls~\cite{zintgraf2017visualizing}. Another potential shortcoming is that given the non-linear nature of deep neural network models, the outputs of these methods are specific to ``the exact subset and modification" of the input features, and are not generalizable to all possible perturbations~\cite{kapishnikov2019xrai}.

\subsection{Contrastive Methods}

A relatively new category of model explanations, contrastive methods go beyond offering explanations for why a certain class was predicted by a model by also providing explanations for why a certain other class was not, something that is ubiquitous in human decision making and explanations. Such explanations are especially important for the assessing the feasibility of deploying automated prediction algorithms in high stakes domains such as medicine, autonomous driving, and criminology. Given the nascence of the field, there are very few methods for contrastive explanations and we review them in this section.

The first contrastive method, called contrastive explanation method (CEM) proposed by Dhurandhar et al.~\cite{dhurandhar2018explanations} used the existing notion of pertinent positive and pertinent negative factors in visual explanations~\cite{herman2016visual}. They used CEM to highlight these factors in any black-box model's predictions. Pertinent positives are features whose presence in the input is ``minimally sufficient" to explain a model's decisions and CEM finds them using an optimization problem which solves for an interpretable perturbation of the input such that this perturbation, when removed from the input, has the same class prediction as the original input. Pertinent negative features are those whose absence is necessary for a class prediction, and they are calculated similarly by optimizing for an interpretable perturbation which when removed from the input, leads to the largest drop in the predicted score for a class.

Another contrastive explanation method called CDeepEx was proposed by Feghahati et al.~\cite{feghahati2019cdeepex} which, unlike CEM, needs access to the model parameters and the model's latent space, but is considerably faster during inference. For seeking a model's explanation for why a certain class was not predicted, CDeepEx learns an adversarial network using the same dataset over which the original classifier is trained on, and for a given input, localizes it in the latent space and finds the smallest distance in the latent space for which the new point corresponds to the class label the that was not predicted. This leads to explanations which are in the same space as the inputs and are ``maximally ambiguous" between the predicted class and the particular class that was not predicted.

More recently, Prabhushankar et al.~\cite{prabhushankar2020contrastive} proposed an approach for obtaining contrastive explanations as an add-on atop existing gradient-based attribution methods. In particular, given a deep neural network model, the predicted class and a class that was not predicted, they use the derivative of the loss (cross-entropy loss and mean-squared error for classification and regression tasks respectively) between the two classes as the contrast. This is then backpopagated to through the network and the authors use Grad-CAM~\cite{selvaraju2017grad} to visualize their explanations as heatmaps. Although they require access to the prediction model, they use only a single backward pass, making the proposed approach considerably faster than previous approaches such as CEM and CDeepEx which require optimization of an objective function.

\section{Evaluating XAI Methods: Shortcomings and Challenges}

Although there has been considerable research in the development of XAI methods for explaining the decisions of prediction models, the evaluation of these methods remains a key challenge. Since saliency maps are not equivalent to localization outputs~\cite{dabkowski2017real}, the majority of these methods are evaluated qualitatively, with the XAI methods' explanations validated \textit{a posteriori}~\cite{fong2019understanding} by assessing their agreement with the region(s) of interest in the image, thus risking human confirmation bias~\cite{adebayo2018sanity}. While measures such as accuracy, fidelity, consistency, certainty, etc. have been proposed to evaluate individual explanations (we direct the interested readers to the survey by Carvalho et al.~\cite{carvalho2019machine}), there is very little research into developing qualitative measures for evaluating XAI methods. Moreover, it has been shown that computer vision algorithms tend to look at different regions of images as compared to humans when performing visual perception tasks~\cite{das2017human}. Therefore, a qualitative assessment of XAI methods might not be sufficient, and reliable quantitative measures which are robust to changes in the computer vision task~\cite{hoffman2018metrics} are desired.

More recently, investigation into the reliability of these saliency methods has shown the inadequacy of some of these methods. Adebayo et al.~\cite{adebayo2018sanity} proposed `sanity checks' based on the randomizations of model parameters and the dataset, and showed that many popular and widely used saliency methods such as guided Grad-CAM~\cite{selvaraju2017grad} and guided backpropagation~\cite{springenberg2014striving} failed these tests, implying that they are independent of both the training data and the model parameters, rendering them inadequate for tasks which involve explaining the relationship between the inputs and the outputs of the trained model (such as finding outliers or debugging the model). Dombrowski et al.~\cite{dombrowski2019explanations} showed that by introducing imperceptible perturbations to the images, it is possible to manipulate the saliency-based explanations for computer vision models without changing the models' predictions. A similar example of misleading explanations was highlighted by Kindermans et al.~\cite{kindermans2019unreliability}, where adding a constant shift to the input showed that atribution methods such as Integrated Gradients~\cite{sundararajan2017axiomatic} and deep Taylor decomposition~\cite{montavon2017explaining} are not input shift invariant and that it is easy to generate ``deceptive" explanations using input shifts. Soon after, Hooker et al.~\cite{hooker2019benchmark} proposed a benchmark for evaluating interpretability methods by first replacing pixels which are deemed to be the most important for the model's accuracy, with random values, for all images in the training and testing data partitions, and then training new models on these modified datasets. They argue that doing so should lead to a sharp degradation in the accuracy, with the accuracy of a feature importance estimator determined by its ability to identify the most important pixels, the removal of which leads to the largest performance drops.

Finally, there have been works arguing in favor of a move towards interpretable models instead of the need to explain the black-box models~\cite{lou2012intelligible,rudin2019stop}, especially for tasks involving high-stakes decisions such as criminal justice, finance, and healthcare. While there appears to be no universally agreed upon formal definition of interpretability, Lipton~\cite{lipton2018mythos} posits that models are interpretable when they offer either transparency (e.g., transparency of each aspect of the model as well as the learning algorithm) or post-hoc explainability (e.g., class activation maps for image classification). The literature surveyed in this review belongs to the second category, because even in scenarios where the deep neural network models' parameters are accessible, these models tend to behave as black-box models because they are ``too complicated for any human to comprehend"~\cite{rudin2019stop}. Rudin~\cite{rudin2019stop} argues that the decision explanations produced by existing methods lend little insight into a deep neural network's often complicated reasoning; for example, saliency maps generated for an image can appear very similar for two very different labels, thus implying that saliency-based methods do not explain anything about the model's decision beyond where in the image the model is looking. Moreover, a trade-off always exists between model interpretability and accuracy is a ``myth", when often in fact, improving the interpretability of a model's predictions leads to improved accuracy.

\section{Conclusion}

The ability to understand decisions made by AI models is of paramount importance and a pre-requisite for the widespread adoption of these models, especially in high stake situations such as autonomous driving, computer-aided diagnosis, and robot-assisted surgery. In this paper, we review the literature on XAI methods in computer vision. In particular, we focus on the attribution-based methods and cover three major categories of approaches: gradient-based, perturbation-based, and contrastive explanations. We discuss the advantages and trade-offs of the methods therein, followed by the challenges involved with the evaluation of these methods, and conclude with an investigation into their reliability.

\footnotesize
\bibliographystyle{plain}
\bibliography{refs}
\end{document}